\documentclass[letterpaper, 10 pt, conference]{ieeeconf}  

\IEEEoverridecommandlockouts                              

\usepackage{graphics} 
\usepackage{amsmath} 

\usepackage{mathtools,amsthm}
\usepackage{amssymb}  
\usepackage{graphicx}
\usepackage[table]{xcolor}
\usepackage{placeins}
\usepackage[font={small}]{caption}
\usepackage{subcaption}
\usepackage{verbatim}
\usepackage{hyperref}
\usepackage{epstopdf}
\usepackage[disable]{todonotes} 

\newcommand{\bts}{{\bar\times^*}}  

\newcommand{\mytheorem}[2]{%
\newtheorem{t#2}{#1}%
\newenvironment{#2}{\begin{t#2}}{\end{t#2}}}

\newcommand{\myremark}[2]{%
\newtheorem{t#2}{#1}[section]%
\newenvironment{#2}{\begin{t#2}}{\end{t#2}}}

\theoremstyle{plain}

\mytheorem{Theorem}{theorem}
\mytheorem{Lemma}{lemma}
\mytheorem{Proposition}{proposition}
\mytheorem{Corollary}{corollary}
\mytheorem{Assumption}{assumption}
\mytheorem{Definition}{definition}
\mytheorem{Property}{property}
\myremark{Remark}{ownremark}

\newcommand{\ls}{\hspace{0em}}     
\newcommand{\rmv}{{\rm v}}      
\newcommand{\rmf}{{\rm f}}      

\title{\LARGE \bf Contact Force and Joint Torque Estimation Using Skin}

\author{Francisco Javier Andrade Chavez$^{1,2}$, Joan Kangro${^2}$, Silvio Traversaro$^{2}$, Francesco Nori$^{2}$ and Daniele Pucci$^{2}$
\thanks{This paper was supported by the FP7 EU projects CoDyCo (No. 600716
ICT 2011.2.1 Cognitive Systems and Robotics) and Koroibot (No. 611909
ICT-2013.2.1 Cognitive Systems and Robotics).}
\thanks{$^{1}$ Francisco Javier Andrade Chavez with CVU 468287 receives support from the National Council of Science and Technology in Mexico, CONACYT}
\thanks{$^{2}$ All authors belong to the Istituto Italiano di Tecnologia, iCub Facility department, Genova, Italy.
       Emails: {\tt\small ing.andrade.francisco@gmail.com},
        {\tt\small joan.kangro@iit.it}, 
         {\tt\small silvio.traversaro@iit.it}, 
          {\tt\small francesco.nori@iit.it}, 
           {\tt\small daniele.pucci@iit.it}}
}

\begin{document}

\maketitle
\thispagestyle{empty}
\pagestyle{empty}

\begin{abstract}
In this paper, we present algorithms to estimate external contact forces and joint torques using only skin, i.e. distributed tactile sensors.
To deal with gaps between the tactile sensors (taxels), we use interpolation techniques. The application of these interpolation techniques allows us to estimate contact forces and joint torques  without the need for expensive force-torque sensors. Validation was performed using the iCub humanoid robot. 

\end{abstract}


\section{INTRODUCTION}

 \todo[inline]{ So what is the structure of intro? 
- general comment
-external contact estimation techniques
-joint torque estimation techniques
-contribution}

At present robots are unable to handle unexpected interactions with their environment, as it could be observed at the  DARPA Robotics  Challenge  Finals  in  June  2015~\cite{Manuelli2016}. To cope with these situations information about both the location of the contact as well as the forces acting on the robot are fundamental to find an appropriate response. As humans, we have skin covering our entire body, which allows  us  to  easily  sense  external contacts and the effect they have on the rest of the body, enabling us to move around even in total darkness. The skin allows to sense contact location, pressure and forces when we interact with our environment. Robots would require to know the location of the contact forces, the value of the external force and how it affects the other parts of the body such as joint torques to replicate this capability.

 Providing force information to the robots increases their operational ability. Contact force estimation enables them to obtain new mechanical characteristics and enables the ability of simplifying many of the control algorithms~\cite{Schneider2006}. Many approaches have been developed to give this capability to robots. They typically separate this problem in two phases: contact location estimation and contact force estimation.
 Sensor fusion techniques have been used to perform contact detection and localization based on hypothesis generated using tactile, force-torque and range sensors~\cite{7041424}. This approach gives mainly the contact location for improving grasping, while estimation of contact forces is neglected. 
 
 A strategy based on the residual concept proposed in~\cite{DeLuca2006}, have been used to estimate contact forces with the aid of extra sensors such as the Kinect to detect the contact location~\cite{Magrini2014}. This approach depends on vision sensors to determine the location of the contacts which could suffer from occlusions during real case scenarios.
 
 Another approach to estimate contact forces has been to use a Kalman filter modeling the forces as spring contacts~\cite{Park2005}. This technique called Active Observer had as objective to estimate the position of the contact point on the robot, while maintaining constant the model of the environment. This technique allows estimation of the contact force, but is limited to the single contact case.
 
 An Extended Kalman filter using the state augmentation method was used to estimate the dynamic pose and internal (body) and external (ground contact) force-torques acting on the individual feet of a bipedal robot fusing haptic (compliant skin), inertial, and force/torque (F/T) measurements~\cite{7353746}. This approach is limited to the foot where the compliant skin is flat.
  
 Given whole-body distributed force/torque and tactile sensors, a strategy using the joint torque estimation from~\cite{Fumagalli2012} can be used to estimate external contact forces~\cite{DelPrete2012}. With this strategy an exact estimation can be given only if there is one contact force-torque per subtree. A linear least-square method is proposed to obtain an approximated solution, when there is more than one contact. 

 Some attempts have been made to calibrate the tactile sensors to estimate contact forces using multiaxis force/torque (F/T) sensors measurements to define a linear regression of the unknown local stiffness. Transformation matrices between the F/T sensor and each tactile element are able to be calculated in the process~\cite{Ciliberto2014}. This technique was conducted in a planar array of tactile sensors manually stimulating each tactile element, disregarding the gaps in between.
 
 Calibration of the skin using vacuum bags to estimate contact forces has also been done~\cite{kangro}.This approach allows to calibrate a whole patch of skin at the same time. Although again the space between the tactile elements is neglected.
 
It has been shown that Joint torque feedback is a fundamental part of force, compliance and impedance control~\cite{944464,4026113}. Strain gauge-based direct sensing of joint torques is one of the main methods used to obtain joint torque feedback~\cite{Wu1980, Luh1983}.

Another popular technique to estimate joint torques is the use of so-called Series Elastic Actuators (SEA)~\cite{Paine2015}. A drawback of SEAs is that, while the additional compliance is convenient for torque measurements, it may be a limitation for higher level controllers not explicitly designed to deal with a compliant system.

An alternative to SEA-based torque estimation is to estimate the torque conveyed by the transmission by measuring directly its deformation, rather than using a specifically designed elastic element. An example is to measure the deformation in the Harmonic Drive to estimate Joint torques~\cite{Zhang2015}.

Using any of these techniques for joint torque estimation, may imply changing the mechanical structure of the joints, which in many cases is not an option for the people working with the robots. Therefore there are many robots that do not include joint torque sensors such as the iCub. A solution has been to add F/T sensors at strategic locations. Installing force sensors on the robots resulted in high maintenance prices, high noise values, soft structure, and complication of the system's dynamic equations~\cite{Katsura2007}.

The main contribution of this paper is to prove that relying only in the skin we can give robots the capability of detecting contact locations, estimate the external contact forces and joint torques. The advantages are that it can be easily integrated on the robot and is cheaper compared to other solutions such as F/T sensors or joint torque sensors. We accomplish this by first improving the accuracy of the tactile sensors by interpolating the pressure values to fill the gaps between tactile elements. Then we estimate the external force. Finally, we use this information to estimate the joint torques of the robot. 

To the best of our knowledge this is the first time a whole body distributed tactile sensor has been effectively used to solve the problem of detecting contact locations, estimate contact forces and estimate joint torques. 

The proposed techniques are validated using the humanoid robot iCub.

\section{BACKGROUND}
\subsection{Notation}

\todo[inline]{traversaro: There are some inconsistency in the notation, but I can through them myself.}

The following notation is used throughout the paper. 
See~\cite{traversaro2016multibody,silvioThesis, saccon2017centroidal} for more details on the notation. 
\begin{itemize}

 \item The Euclidean norm of either a vector or a matrix of real numbers is denoted by $\left\| \cdot \right\|$.

\item Coordinate frames are indicated with capital letters, such as $B$ and $L$. $A$ indicates an inertial frame. 
\item ${}^A R_B \in \mathbb{R}^{3 \times 3}$ is the 3D rotation matrix from $B$ to $A$, and $\ls^A o_B \in \mathbb{R}^3$ are the coordinates of the origin of frame $B$ expressed in frame $A$.
\item Given $u,v \in \mathbb{R}^3$, $u^\wedge \in \mathbb{R}^{3\times3}$ denotes the skew-symmetric matrix-valued operator associated with the cross product in 
  $\mathbb{R}^3$, such that $u^\wedge v = u \times v$.
\item $\ls^A \omega_{A, B}$, with $\ls^A \omega_{A, B}^\wedge = \ls^A \dot{R}_B \ls^A R_B^\top $ is the angular velocity of frame $B$ with respect to the frame $A$ expressed in frame $A$.
\item 
$\ls^C \rmv_{A,B} =
\begin{bsmallmatrix}
\ls^C v_{A,B} \\
\ls^C \omega_{A,B}
\end{bsmallmatrix}
=
\begin{bsmallmatrix}
    \ls^C R_A \ls^A \dot{o}_B \\
    \ls^C R_A \ls^A \omega_{A,B} 
\end{bsmallmatrix}
$
is the 6D velocity of the frame $B$ with respect to the reference frame $A$, expressed in frame $C$
\item $\ls_B \rmf = \begin{bsmallmatrix} \ls_B f \\ \ls_B \mu \end{bsmallmatrix}$ are the coordinates of the 6D force $\rmf$ expressed in the $B$ frame, 
\item $\ls^C \rmv_{A,B}\bts 
= 
\begin{bsmallmatrix}
\ls^C\omega_{A,B}^\wedge & 
0_{3 \times 3} \\
\ls^Cv_{A,B}^\wedge & 
\ls^C\omega_{A,B}^\wedge
\end{bsmallmatrix}
$
is the matrix representation of the dual cross product 
\item 
$
\ls_AX^B
= 
\begin{bsmallmatrix}
\ls^A R_B & 0_{3 \times 3} \\
\ls^A o_B^\wedge \ls^A R_B & \ls^A R_B 
\end{bsmallmatrix}
$  is the 6D force transformation   from $B$ to $A$

\item A \emph{multibody} system is a couple of a set $\mathfrak{L}$ of $n_L$ rigid bodies --called \emph{links}-- interconnected by $n_J$ mechanisms --called \emph{joints}-- constraining the relative motion of a pair of links. $\mathfrak{J}$ is the set of joints, represented as set of the two links interconnected by the joint. Every body $B$ is associated with a frame $B$ rigidly attached to it. 
\item 
$
\ls_{B} \mathbb{M}_B = 
\begin{bsmallmatrix}
      m 1_{3 \times 3}       &  m \, \ls^B c^{\wedge} \\ 
     -m \, \ls^B c^{\wedge}  &  \ls_B \mathbb{I}_B
  \end{bsmallmatrix}
$
is the inertia tensor of body $B$ expressed with respect to frame $B$,
where $m$ is the body mass, $\ls^B c$ are the coordinates of the center of mass in frame $B$ and $\ls_B \mathbb{I}_B$ is the 3D inertia matrix of the rigid body, expressed with the orientation of frame $B$ and with respect to the frame $B$ origin. 
\item $\tau_{\{A,B\}}$ is the joint torque of the joint connecting link A and B.
\item $\left< s, p \right>$ is the dot product between vectors s and p.
\item $
\ls^B \alpha_{A, B}^g = 
\begin{bsmallmatrix}
\ls^B R_A ( \ls^A \dot{o}_B - g ) \\
\ls^B \dot{\omega}_{A, B}
\end{bsmallmatrix}
$ is the \emph{sensor proper} acceleration of the frame $B$ w.r.t. to the frame $A$~\cite[Chapter 3]{silvioThesis}, where $\ls^A g$ is the gravitational acceleration in the inertial frame. 

\end{itemize}

\subsection{Distributed Tactile Sensors}
The distributed tactile sensors (artificial skin) are used in many robotic applications to get the feedback about the environment. 
It usually consists of an array of discrete sensors that allow to detect contacts. There are skins with various characteristics, but this paper assumes that the skin available in the robot is able to measure the pressure applied to it. It also assumes that sensor locations and orientations with respect to the robot frames are known.

\subsection{Joint torque estimation}
\label{section:Joint Background}
\begin{figure}
\vspace*{0.2cm}
\centering
\includegraphics[width=0.25\paperwidth]{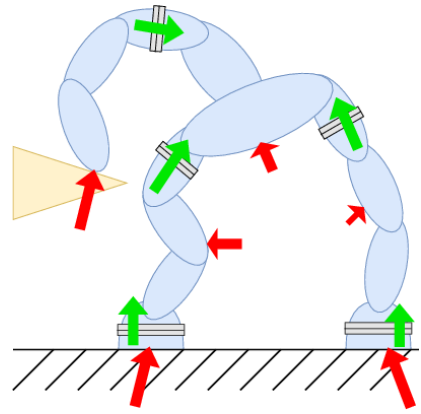}
\caption{Example of a multibody system with internal six-axis force-torque sensors. Measured force-torques are indicated in green, while unknown contact force-torques are drawn in red. There are $n = 5$ force-torque sensor in the system, that is then decomposed in $n+1 = 6$ submodels for external force-torque estimation.}
\label{fig:extFTsMultiBody}
\end{figure}

\begin{figure}
\vspace*{0.2cm}
\includegraphics[width=0.40 \paperwidth]{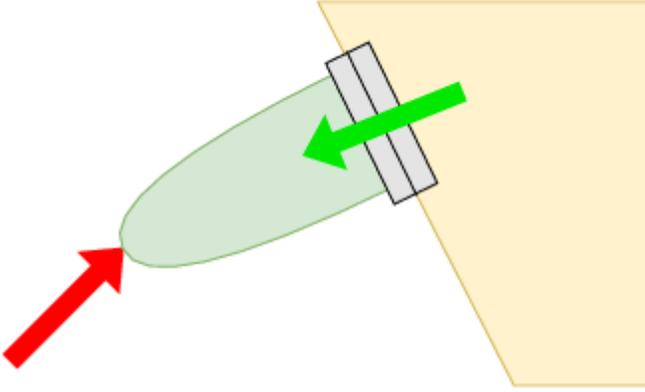}
\caption{Graphical representation of equation \eqref{eq:singleBodyForce}.}
\label{fig:extFTsSingleBody}
\end{figure}

What follows is a description of the theoretical framework proposed in~\cite{Fumagalli2012,DelPrete2012} for the estimation of external force and joint torques on chains, later extended for the whole-body case in~\cite{silvioThesis}.
The proposed  algorithm consists in cutting the floating-base tree at the level of the (embedded) F/T sensors obtaining multiple subtrees as in Fig.~\ref{fig:extFTsMultiBody}. Each subtree is considered an independent articulated floating-base structure governed by the Newton-Euler dynamic equations. The F/T sensor, gives a direct measurement of one specific external force-torque acting on  the structure (green arrows in Fig.~\ref{fig:extFTsMultiBody}). Other external force-torques (red arrows in Fig.~\ref{fig:extFTsMultiBody} and Fig.~ \ref{fig:extFTsSingleBody}) are estimated with the procedure described below. 

\subsubsection{Method for estimating external force-torques} \label{sec:externalforce-torques} 
In the simple case of one body, assuming a perfect model and the knowledge of the inertial parameters of the body,  measurements of the sensor proper acceleration of body $B$ ($\alpha^g_{A,B}$) , the angular velocity of the link $B$ in the $B$ frame ($^B \omega_{A,B}$) and force-torque sensor measurements ($\ls_B \rmf^s$) at a given instant, we can estimate the external force-torque $\ls_B \rmf^x$ by writing the Newton-Euler equations for body $B$:
\begin{equation}
\label{eq:singleBodyForce}
    \ls_B \rmf^x = \ls_B\mathbb{M}_B \alpha^g_{A,B} + \begin{bmatrix} 
0_{3\times1} \\
^B \omega_{A,B} 
\end{bmatrix}
\bar{\times}^*
_B \mathbb{M}_B
\begin{bmatrix} 
0_{3\times1} \\
^B \omega_{A,B} 
\end{bmatrix} - \ls_B \rmf^s .
\end{equation}

Conveniently, in this formula the term $\mathbb{M} \alpha^g_{A,B} + \begin{bmatrix} 
0_{3\times1} \\
^B \omega_{A,B} 
\end{bmatrix}
\bar{\times}^*
_B \mathbb{M}_B
\begin{bmatrix} 
0_{3\times1} \\
^B \omega_{A,B} 
\end{bmatrix}$ is the only one that depends  on acceleration, velocity and the inertial parameters of the body. For convenience, we will indicate this term as:
\begin{equation}
_B \phi_B( ^B \alpha^g_{A,B}, ^B \omega_{A,B})\! := _B\! \mathbb{M}_B \alpha^g_{A,B} +\! \begin{bmatrix} 
0_{3\times1} \\
\omega_{A,B} 
\end{bmatrix}
\bar{\times}^*
_B \mathbb{M}_B \!\!
\begin{bmatrix} 
0_{3\times1} \\
^B \omega_{A,B} 
\end{bmatrix}.
\end{equation}

From here on we will omit the dependency on the \emph{proper sensor acceleration} and on the \emph{body angular velocity} and simply indicate this as $_B \phi_B$. An interpretation for the physical meaning of $_B \phi_B$ is the sum of all the force-torque acting on body, both the external ones and the one due to interaction with the other bodies in the system, minus the gravitational force-torque. Even if this term does not include the force-torque due to gravity, to simplify the nomenclature in the following will call it \emph{net force-torque} acting on the body $B$.

When considering the case of a multibody system, for each link $L \in \mathfrak{L}_{sm}$ we indicate with $\beth_{sm}(L)$ the set of links that are connected with $L$ in the full model, but that belong to a different submodel, i.e.: 
\begin{equation}
\beth_{sm}{(L)} := \{ D \in \mathfrak{L} \  | \  \{L,D\} \in \mathfrak{J} \land D \notin  \mathfrak{L}_{sm}\} .
\end{equation}

For the multibody case we express the net force-torque as:
\begin{align}
\label{eq:NEforEstimationSubModel}
    \sum_{L \in \mathfrak{L}_{sm}}  {}_B X^L \, _L \phi_L =& \sum_{L \in (\mathfrak{C} \cap \mathfrak{L}_{sm})} \, \ls_B X^L\, \ls_L \rmf^x_L \\ +&  \sum_{L \in \mathfrak{L}_{sm}} \sum_{D \in \beth_{sm}(L)} {}_B X^D\, \ls_D \rmf_{D,L} .
\end{align}
Where $\ls_L \rmf^x_L$ is the external force-torque of link L expressed in link L frame, $\ls_D \rmf_{D,L}$ is the force-torque that link $D$ exerts on link $L$ as seen by the F/T sensor in between both links and $\mathfrak{C}  \subseteq \mathfrak{L}$ is the subset of the links where external force-torques are acting . 
Noting that in (\ref{eq:singleBodyForce}) and in (\ref{eq:NEforEstimationSubModel}) the only unknowns are the contact force-torques, the estimation problem may be solved rewriting these equations in the matrix form $A{x}={b}$, where ${x=\sum_{L \in (\mathfrak{C} \cap \mathfrak{L}_{sm})} {}_L f^x_L}\in \mathbb{R}^{u}$ contains all the $u$ contact unknowns, whereas $A\in \mathbb{R}
^{6\times u}$ and ${b}\in \mathbb{R}^{6}$ are completely determined. 

Using the previous equations we take into consideration the following three types of possible contacts:
 \begin{itemize}
     \item pure force-torque ($\ls_L{\rmf}^x$, 6-dimensional unknown vector corresponding to force and torque expressed in the link frame of contact $L$)
     \item pure force ($f^x$, 3-dimensional unknown vector corresponding to a pure force and no torque)
     \item force norm ($\|{f}^x\|$, one-dimensional unknown assuming the pure force to be orthogonal to the contact surface)
 \end{itemize}
 The matrix $A$ is built by adding columns for each contact according to its type. The columns associated to pure force-torques ($A_w$), pure forces ($A_f$) and pure force norm ($A_n$) are the following:

\begin{IEEEeqnarray*}{rCl}
A_w =& \left[  \ls_B X^L \right],
\quad \\
A_f =& \left[ \begin{array}{c}
\ls^B R_L\\
0_{3\times3}\\
\end{array} \right],
\quad \\
A_n =& \left[  _B X^L \right] \left[ \begin{array}{c}
\hat{{u}}^x\\
0_{3\times1}\\
\end{array} \right].
\end{IEEEeqnarray*}

\noindent
where $B$ is a common frame which in this case was selected as the base of the subtree and $\hat{{u}}^x$ is the unit normal vector of the contact force-torque. The matrix $A$ mainly depends on the contact location sensed by the skin. The 6 dimensional vector ${b}$ is defined from \ref{eq:NEforEstimationSubModel} in the following way: 
\begin{equation}
{b} = 
 \sum_{L \in \mathfrak{L}_{sm}} {}_B X^L \, _L \phi_L  - \sum_{L \in \mathfrak{L}_{sm}} \sum_{D \in \beth_{sm}(L)} {}_B X^D\, _D f_{D,L} .
\end{equation}

\noindent

The vector ${b}$ depends on kinematic quantities which can be derived for the whole-body distributed gyros, accelerometers, encoders and the F/T sensors. Details on how to estimate this quantities are provide by~\cite{Fumagalli2012}. Once $A$ and ${b}$ have been computed, we can solve the equation $A {x} = {b}$ for estimating external force-torques. 

When only a single contact acts on the subtree the associated force-torque has a unique solution,since there are six unknowns for a system of six equations. An exact characterization of the external and internal force-torques can be obtained if there exists only one contact force-torque per each of the subtrees obtained by the body structure partition induced by the F/T sensor locations (see Fig.~\ref{fig:sensors}). In all other cases, an exact estimate cannot be obtained but a reasonable estimate of all the contact force-torques can still be obtained. The adopted solution  consists in computing the minimum norm ${x}^*$ that minimizes the square error residual:
\begin{equation*}
{x}^* = A^{\dagger}{b}
\end{equation*}
where $A^\dagger$ is the Moore-Penrose pseudo-inverse of $A$ \cite{DelPrete2012}.

\subsubsection{Method for estimating internal torques.} \label{sec:internalTorques} The joint torques estimation is crucial because this quantity is directly related to the motors that are actuating those joints, and the input to the motors is in the end the ultimate control input available in robots. Once an estimate of  external forces is obtained with the method described in Section \ref{sec:externalforce-torques}, internal force-torques can also be estimated with a standard Recursive Newton-Euler Algorithm (RNEA). The torque $\tau_{\{E,F\}}$ of the joint comes from the projection of the joint force-torque on the joint motion subspace :
\begin{equation}
\label{eq:jointTorqueFromInternalSimple}
\tau_{\{E,F\}} = \left< \rmf^F \mathrm{s}_{E,F}, \ls_F f_{E,F} \right> = \left< ^E \mathrm{s}_{F,E}, \ls_E \rmf_{F,E} \right> ,
\end{equation}

\begin{IEEEeqnarray}{rCl}
\label{eq:internalforce-torqueFromNetAndExternal}
\IEEEyesnumber 
 \ls_F \rmf_{E,F}  &=& - _E f_{F,E}, \IEEEyessubnumber \\ 
\ls_F \rmf_{E,F} &=& \sum_{L \in \gamma_E(F)} {}_F X^L  \left( _L\phi_L + \ls_L \rmf^x_L \right), \IEEEyessubnumber \label{eq:otherExpressionForInternalFT} \\
_E f_{F,E} &=& \sum_{L \in \gamma_F(E)} {}_E X^L \left( _L\phi_L + _L f^x_L  \right), \IEEEyessubnumber \label{eq:oneExpressionForInternalFT} 
\end{IEEEeqnarray}
where $\gamma_E(F)$ is the set of the links belonging to the subtree starting at link $F$, given $E$ as a base link, $^E \mathrm{s}_{F,E}$ is the joint motion subspace.

\section{METHODOLOGY}
\label{sec:method}
\subsection{Problem Statement}
 Assuming a contact is sensed by the skin and its location is known, we need to estimate the external force and joint torques of the robot. We achieve this by improving the contact force estimation of the skin and use this information to estimate the joint torques.

\subsection{Contact Force Estimation From The Skin}

The tactile sensors are able to measure the pressure applied to each one of them individually. However, the sensors are positioned on the surface as an array of discrete taxels with gaps between them. In order to compensate for this shortcoming we can interpolate the pressure values between the sensors where the pressure cannot be explicitly measured. 

Every point on the skin covering a given link can be represented by a pair of surface coordinates, that we refer as the couple $(u,v) \in [u_1, u_2
] \times [v_1, v_2]$.

The method used for interpolation consists of the following steps: 

\begin{enumerate}
  \item The locations of the sensors (on u-v plane) and their pressure values are gathered.
  \item The sensors are modelled as a circle (with appropriate area) labelled with a certain number of data points. The z axis corresponds to the pressure value of a specific sensor. The pressure is assumed to be constant over the area of the sensor, therefore all the data points of a specific sensor have the same z axis value.
  \item The trilinear interpolation based on a Delaunay triangulation is used to interpolate the pressure field between the data points~\cite{interpolation_method}.
\end{enumerate}

\begin{figure*}[t!]
    \centering
    \begin{subfigure}[t]{0.32\textwidth}
        \centering
        \includegraphics[width=\textwidth]{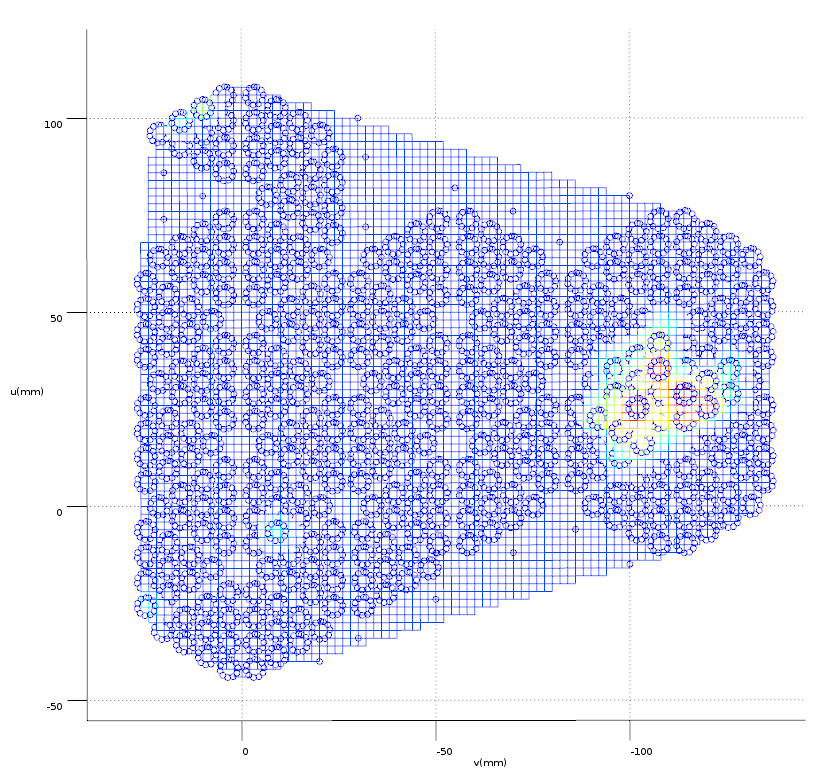}
        \caption{Top view}
    \end{subfigure}%
      ~ 
    \begin{subfigure}[t]{0.32\textwidth}
        \centering
        \includegraphics[width=\textwidth]{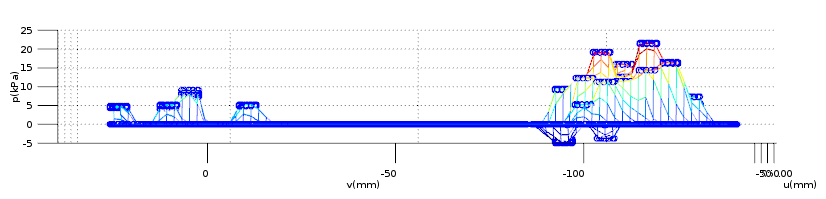}
        \caption{Side view}
    \end{subfigure}
    ~ 
    \begin{subfigure}[t]{0.32\textwidth}
        \centering
        \includegraphics[width=\textwidth]{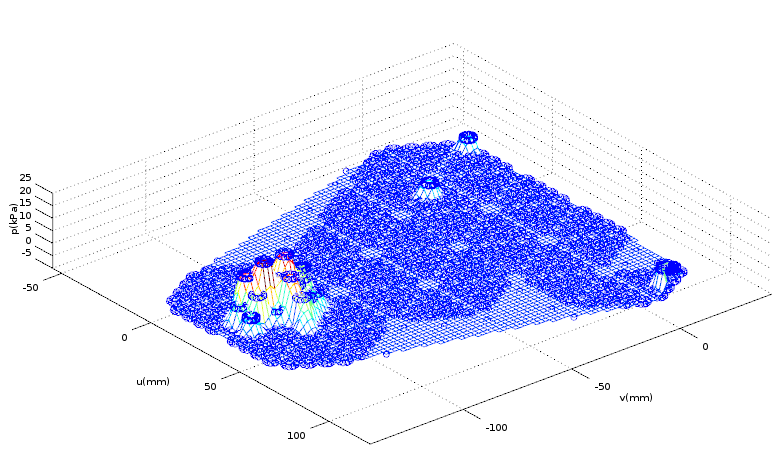}
        \caption{Three-Point Perspective}
    \end{subfigure}
  \caption{Pressure field of a skin patch while 1 kg is applied.}
       \label{fig:skinPerspectives}
\end{figure*}

The output from the interpolation allows us to define the pressure field $ p(u,v) $. An example of the pressure field while a 1 kg mass is put on the skin is shown on figure \ref{fig:skinPerspectives}.

The 3D positions of all the sensors are known but there is no information about the surface between the sensors. Therefore, the surface has to be interpolated between the known values. The positions corresponding to u-v field can be divided into 3 separate interpolation problems, one for each axis. The trilinear interpolation (same as mentioned above) allows us to define the interpolated field of each axis of the position vectors, ie ${x}(u,v)$, ${y}(u,v)$ and ${z}(u,v)$.

The position vector corresponding to a location on u-v plane can be expressed as follows:

\begin{equation}
{r}(u,v)= {x}(u,v) e_1 + {y}(u,v) e_2 + {z}(u,v) e_3 .
\end{equation}

The normal vectors of all the sensors are known but there is no information about the surface normals between the sensors. Therefore, the normals have to be interpolated between the known values. The normals corresponding to u-v field can also be divided into 3 separate interpolation problems, one for each axis. The trilinear interpolation (same as mentioned above) allows us to define the interpolated field of each axis of the unit vectors, ie ${n_x}(u,v)$, ${n_y}(u,v)$ and ${n_z}(u,v)$, with the actual normal $\hat{n
}(u, v)$ given by

\begin{equation}
\hat{{n}}(u,v)=\frac{{n_x}(u,v) e_1 +{n_y}(u,v) e_2 +{n_z}(u,v) e_3}{\mid\mid{n_x}^2(u,v)+{n_y}^2(u,v)+{n_z}^2(u,v)\mid\mid}.
\end{equation}

Assuming that $\left| \frac{\partial r}{\partial u} \times \frac{\partial r}{\partial v} \right| \approx 1$  the total force vector can be found as:

\begin{equation}
f=\int_{v1}^{v2} \int_{u1}^{u2} p(u,v) \hat{{n}}(u,v) du dv, \label{eq:totalForce}
\end{equation} 

while the total torque vector can be found as follows:

\begin{equation}
\mu=\int_{v1}^{v2} \int_{u1}^{u2} ((p(u,v) \hat{{n}}(u,v)) \times {r}(u,v)) du dv.
\end{equation}

\subsection{Adding Known External Force-torques To The Joint Torque Estimation Scheme}
\label{sec:newEstimation}

To consider the effect of having the knowledge of an external force-torque at a known location ($_Kf^k_{K} $) it is necessary to extend the current framework detailed in \ref{section:Joint Background} to include the new type of contact. This is achieved by adding the characteristics of this contact to all parts of the equation $A{x}={b}$ and equations \eqref{eq:otherExpressionForInternalFT}\eqref{eq:oneExpressionForInternalFT}.

In the case of the $A$ matrix the $4_{th}$ case would be a $6 \times 0$ matrix:
\begin{equation}
A_k = 0_{6 \times 0}
\end{equation}
and for the $b$ term the equation would be:

\begin{equation}
{b} = \sum_{L \in \mathfrak{L}_{sm}} {}_B X^L \, _L \phi_L  - _Bf^{tot}_B ,
 \end{equation}
 
 where
 \begin{equation}
 \ls_B \rmf^{tot}= \left(\sum_{L \in \mathfrak{L}_{sm}} \sum_{D \in \beth_{sm}(L)} \!_B X^D\, \ls_D \rmf_{D,L} - \sum_{K \in \mathfrak{K}_{sm}} {}_B X^K\, _Kf^k_{K} \right).
\end{equation}
$\mathfrak{K}_{L}$ is the set of force-torque contacts estimated by the skin that belong to a given link $L$.\\
For the joint torque estimation we have to add the known force-torque to the estimated force-torques:

\begin{align}
\label{eq:newInternalForce-torque}
\IEEEyesnumber 
 {}_F \rmf_{E,F}  =& - {}_E \rmf_{F,E} \\ 
{}_F \rmf_{E,F} =& \sum_{L \in \gamma_E(F)} {}_F X^L\!  \left( {}_L\phi_L + {}_L \rmf^x_L +  \sum_{K \in \mathfrak{K}_{L}} {}_L X^K\, \ls_K \rmf^k_{K}\right) \label{eq:newTorqueEF} ,\\
{}_E \rmf_{F,E} =& \sum_{L \in \gamma_F(E)} {}_E X^L \left( _L\phi_L + {}_L \rmf^x_L  + \sum_{K \in \mathfrak{K}_{L}} {}_L X^K\, {}_K \rmf^k_{K}\right).  \label{eq:newTorqueFE} 
\end{align}

These changes allow the robot to determine multiple external contact forces correctly as long as most of the contacts happen in the areas covered by skin. 

\section{EXPERIMENTS} 
\subsection{Experimental Platform}
\begin{figure}

  \begin{minipage}[c]{0.48\textwidth}
    \begin{subfigure}[t]{0.31\textwidth}
        \centering
        \includegraphics[width=\textwidth]{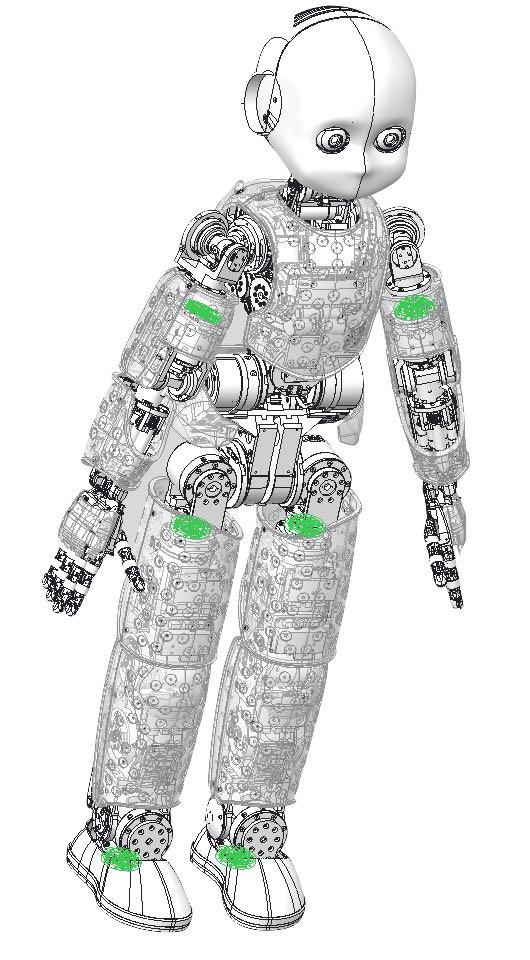}
        \caption{ The six axis F/T sensors location on the iCub.}
    \end{subfigure}%
      ~ 
    \begin{subfigure}[t]{0.65\textwidth}
        \centering
        \includegraphics[width=\textwidth]{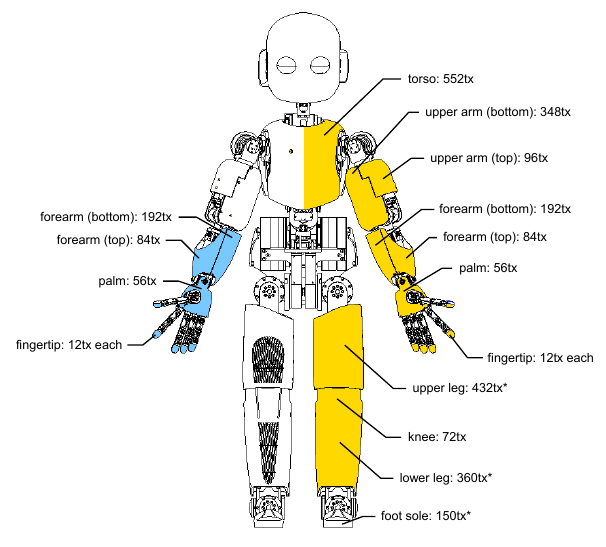}
        \caption{iCub skin distribution on the iCub.}
    \end{subfigure}
  \end{minipage}\hfill
  \caption{iCub sensors used to estimate external sensors}
  \label{fig:sensors}
  \end{figure}

Experiments have been performed on the 53 DOF robot iCub. Six custom-made six axes F/T sensors~\cite{IITsensors}, one per ankle, leg and arm, are placed as shown in Fig. \ref{fig:sensors}.  The distribution of the skin on the robot can be observed in Fig.  \ref{fig:sensors}. For the experiments a set of calibration masses (200gr, 500gr, 1kg and 2kg) were used. The weights are positioned either directly on the right lower leg of the iCub or hanging from the leg with a cloth stripe as shown in Fig. \ref{fig:experiments}.

\todo[inline]{Kangro: I took this subsubsection off as it made it look really weird for some reason}

The force torque sensors mounted on the iCub are six axis force torque sensors with silicon strain gauge technology. This technology bases its measurements in the changes of the resistance according to small deformations of the material. The sensor is designed such that the resulting deformation in the sensor's structure is inside the linear section of the material for the specified range.

The skin of iCub~\cite{Cannata2008} is an array of compliant distributed pressure sensors composed of the flexible printed circuit boards (FPCB) covered by a layer of elastic fabric further enveloped by a thin conductive layer. The FPCB is composed of triangular modules of 10 taxels each, which act as capacitive sensors,  plus  two  temperature  sensors  for  drift  compensation. The tactile sensors have a measurable pressure range up to 180 kPa~\cite{bartolozzi2016robots}. Each single taxel has 8 bits of resolution. The skin of iCub is calibrated using the vacuum bags, that are wrapped around the skin. The air is sucked from the bag in order to reduce the pressure within the bag. This creates a uniform pressure distribution on the skin's surface that enables us to a generate mathematical model for each sensor that relates the capacitance value to the applied pressure~\cite{kangro}. Therefore, we are able to know the pressure that is applied to each separate sensor in the array. 

In this paper, we focus on the right lower leg of the robot and more specifically in the knee joint. The sensors involved in the estimation of the joint torque at the knee are the F/T sensors in the right leg and the skin patch in the right lower leg of the iCub that has a total of 380 discrete sensors.

\begin{figure}[ht!]
    \centering
    \begin{subfigure}[t]{0.15\textwidth}
        \centering
        \includegraphics[width=\textwidth]{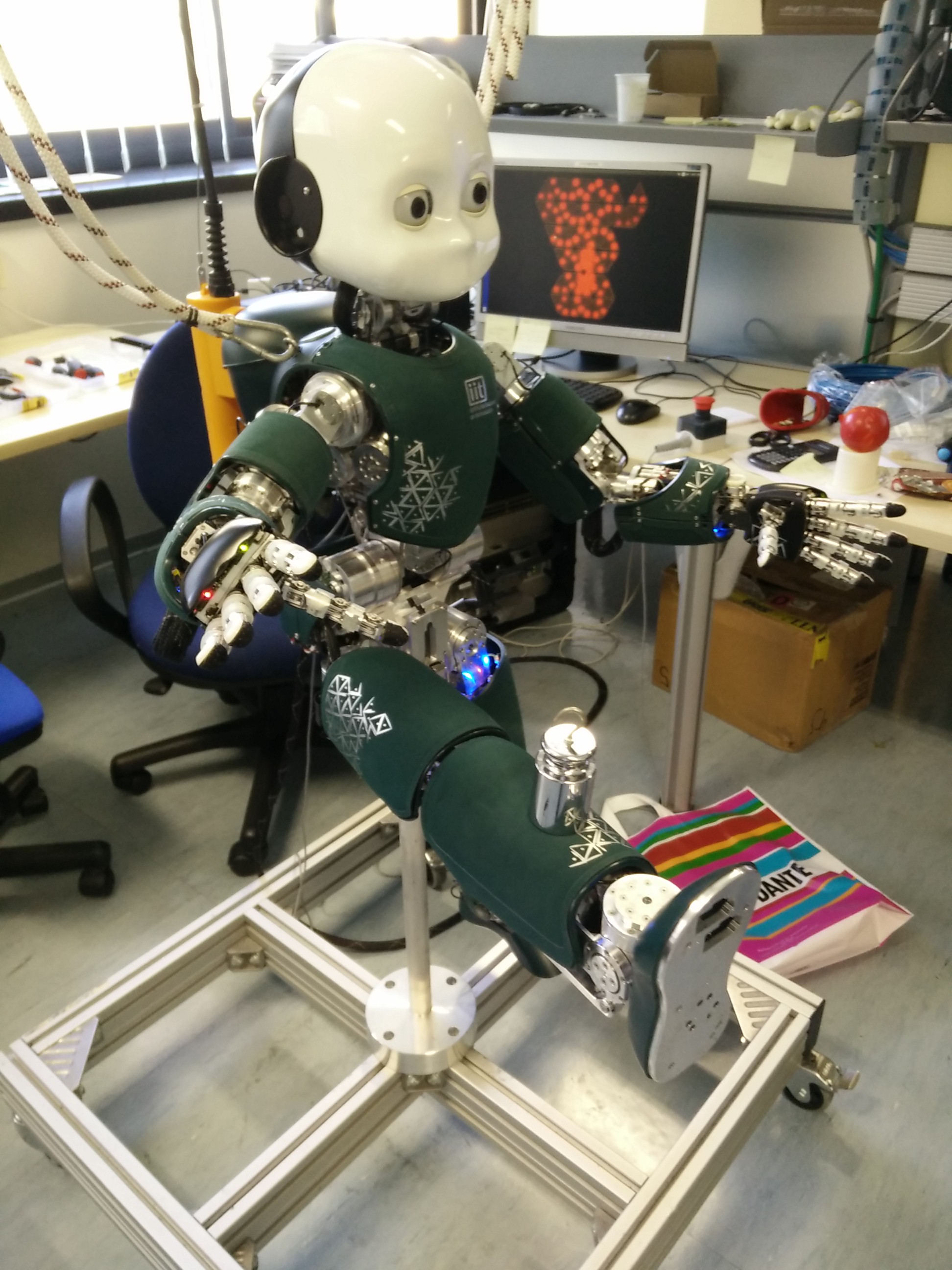}
        \caption{Adding 1kg mass on top of right lower leg}
        \label{fig:onTop}
    \end{subfigure}%
      ~ 
    \begin{subfigure}[t]{0.15\textwidth}
        \centering
        \includegraphics[width=\textwidth]{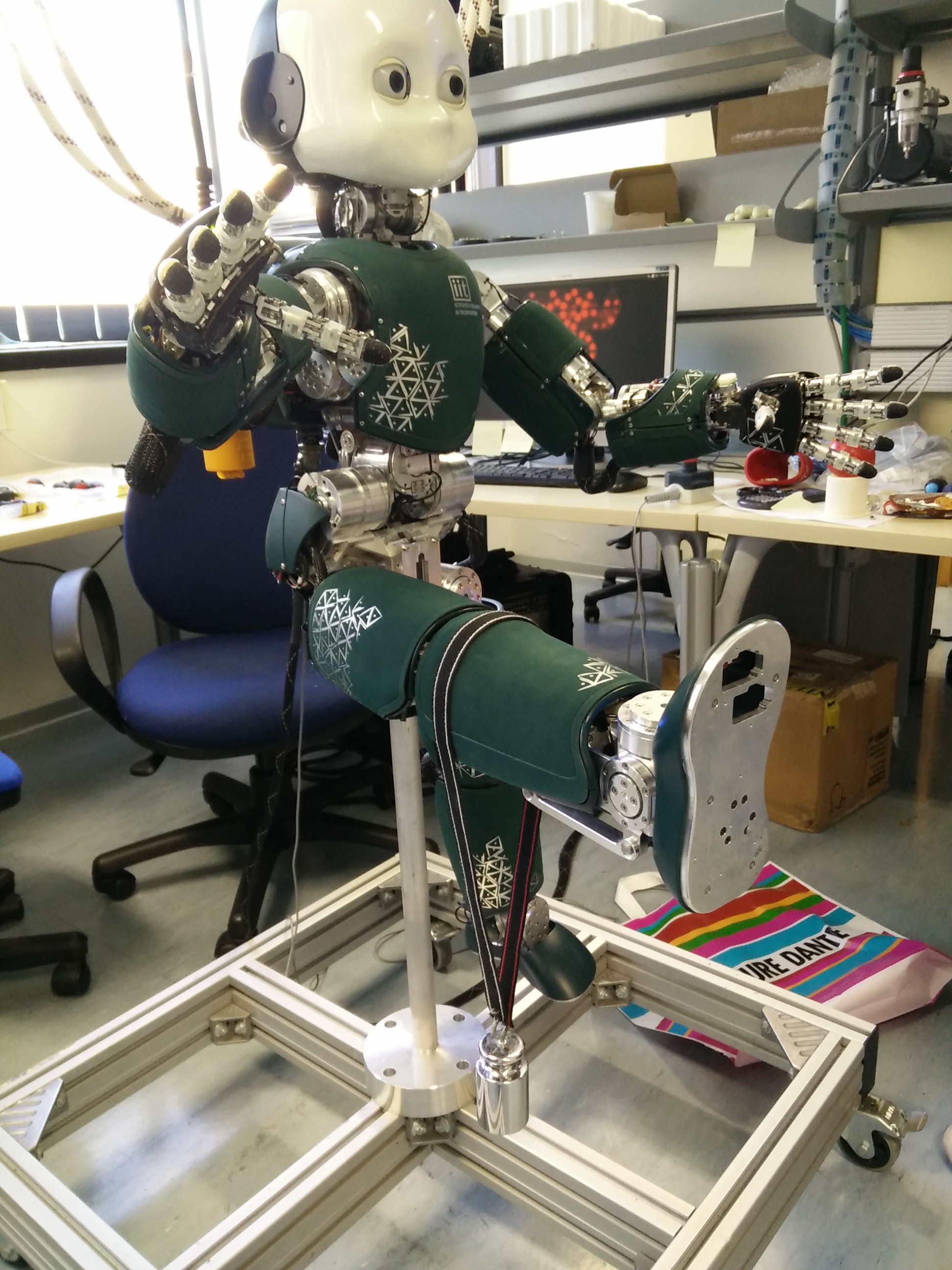}
        \caption{Hanging 1kg mass from right lower leg}
        \label{fig:hanging}
    \end{subfigure}
    ~ 
    \begin{subfigure}[t]{0.15\textwidth}
        \centering
        \includegraphics[width=\textwidth]{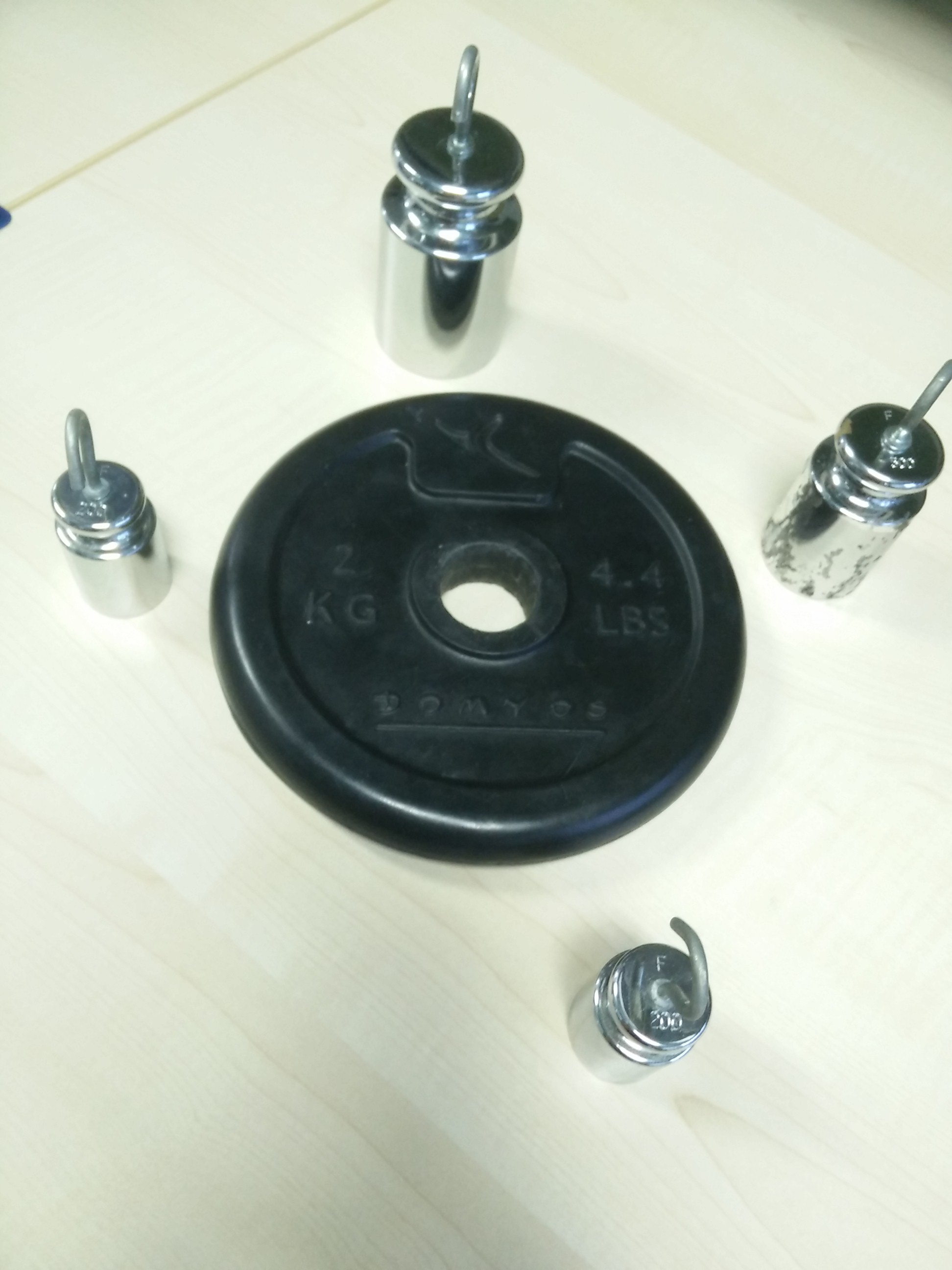}
        \caption{Set of calibration masses used in the experiments} \label{fig:masses}
    \end{subfigure}
\end{figure}
\label{fig:experiments}

\subsection{ Assumptions}

\begin{itemize}
    \item The inertial parameters of the robot are known.
    \item The position of the taxels is known and included in the urdf model of the robot.
    \item The robot skin has been previously calibrated (up to 50kPa) using vacuum bags, using the technique described in~\cite{kangro}.
    \item The F/T sensors where calibrated \textit{in situ} to boost performance of the sensors, using the technique described in~\cite{insituFTcalibration}.
\end{itemize}

\subsection{Experiment Description} 
There were mainly 2 locations in which the weights were applied. When the calibration weights are placed on the right lower leg in a position close to the ankle the distance from the knee is around 12 $\sim$ 13cm. On the other hand when hanging from the lower leg near the knee the distance is around 4 $\sim$ 5cm. The torques are estimated with respect to the frame of the joint in the knee. 

The external force-torques and joint torques obtained using the F/T measurements are estimated using the methods described in \ref{section:Joint Background}. When using the skin the external force-torques and joint torques are directly measured using the calibrated skin and are included as known external force-torques into the extended estimation scheme described in \ref{sec:newEstimation}. An example of the skin being activated by the contact and its pressure field representation can be seen in Fig. \ref{fig:skinPerspectives}.
 
\subsection{ Validation }

For the validation of the forces the calibrated masses were placed on top of the skin normal to the ground as shown on Fig. \ref{fig:onTop}.
The forces applied by the masses were then compared to the forces calculated with and without using the interpolation method. For comparison we use the magnitude of the contact forces calculated using equation \eqref{eq:totalForce}.

The magnitude of the forces applied on the robot estimated without interpolation assuming that the pressure is uniform over the area of the sensor and all the sensors are covering an equal area can be found with the following equation:

\begin{equation} \label{eq:simplified_equation}
\| f_c \| = \| A \sum\limits_{i=1}^k  p_{i} \hat{n}_i \|
\end{equation}

where $f_c$ is the total contact force, $p_{i}$ is the pressure of a particular sensor, $A$ is the area of the tactile sensor, $\hat{n}_i$ is the normal of the taxel and $k$ is the total amount of taxels. 

For the torque, finding a ground truth of the exact location of the contact is required. This location is estimated from the taxels that are activated by the contact. In this case we will consider proximity to the torques estimated with the F/T sensor as the validation, since these values currently allow the iCub robot to perform dynamic movements such as balancing~\cite{nava2016}.

\subsection{Results}
\label{sec:results}

\subsubsection{Contact Force Estimation Results}

\begin{figure}[t!]
\centering
\includegraphics[width=.5\textwidth]{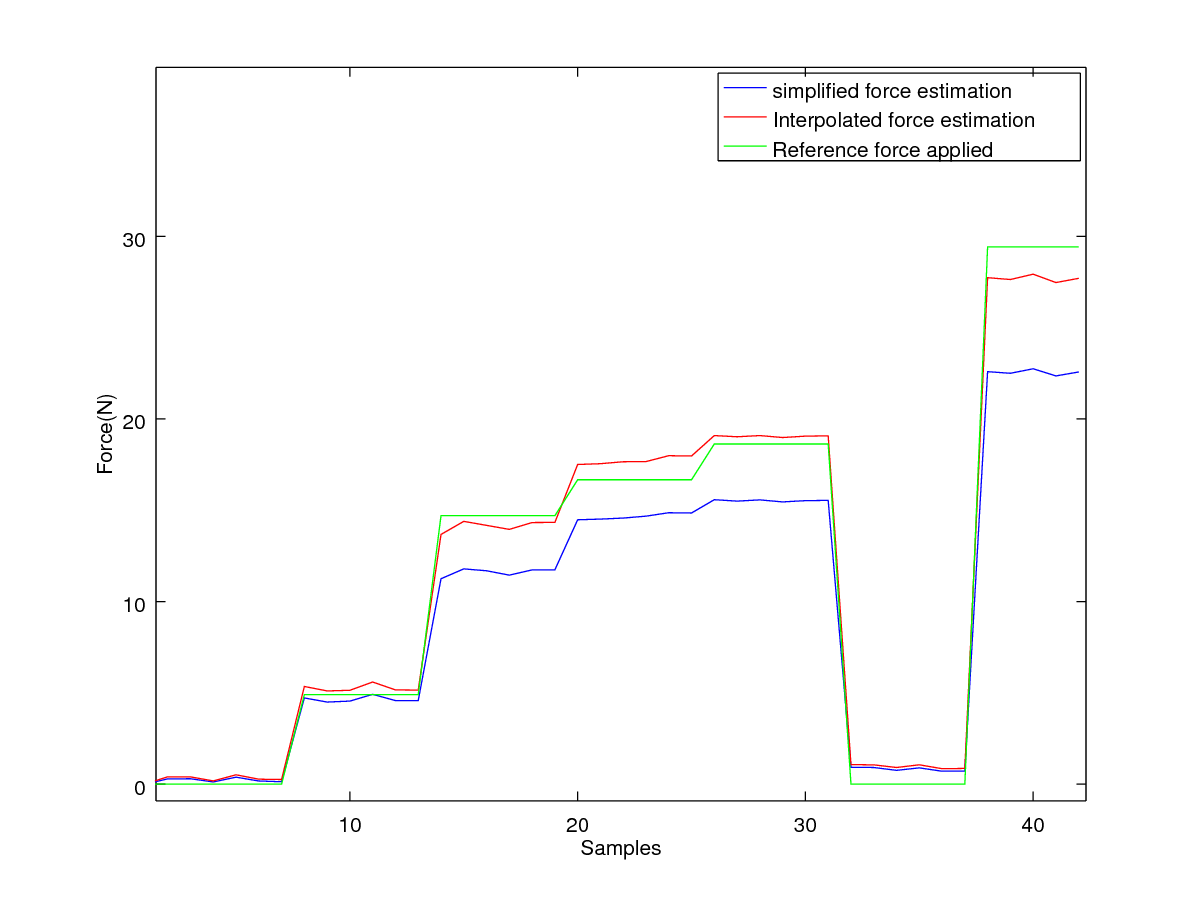}
\caption{Force comparison between reference force applied, interpolated force estimation and simplified force estimation. After every 5-6 samples the program was stopped in order to change the weights applied on the skin. }
\label{fig:force_comparison}
\end{figure}

It can be observed that the estimation without the interpolation underestimates the total force applied. This is due to the fact that some of the force is applied in the areas between the sensors that we cannot measure explicitly. However, interpolation of the pressure field allows us to improve the estimation as can be seen from the graph on Fig.\ref{fig:force_comparison}. The interpolated force (red) calculation, given by the magnitude of Eq.\ref{eq:totalForce}, is compared to the simplified force (blue) calculation, given by Eq. \ref{eq:simplified_equation}. The green line displays the reference force applied on the skin during the experiment. 

\begin{figure*}[ht!]
    \centering
    \begin{subfigure}[t]{0.32\textwidth}
        \centering
        \includegraphics[width=\textwidth]{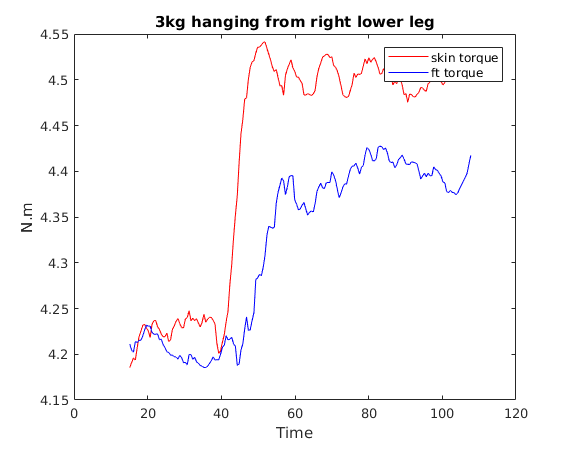}
        \caption{Hanging 3kg mass from right lower leg}
        \label{fig:3kTorque}
    \end{subfigure}
    ~ 
    \begin{subfigure}[t]{0.32\textwidth}
        \centering
        \includegraphics[width=\textwidth]{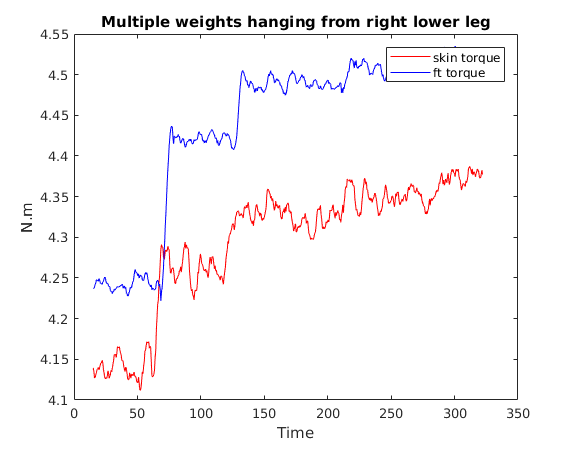}
        \caption{Hanging 1kg, 1.5kg, 1.7kg, 1.9kg consequently} 
        \label{fig:many}
    \end{subfigure}
     ~ 
    \begin{subfigure}[t]{0.32\textwidth}
        \centering
        \includegraphics[width=\textwidth]{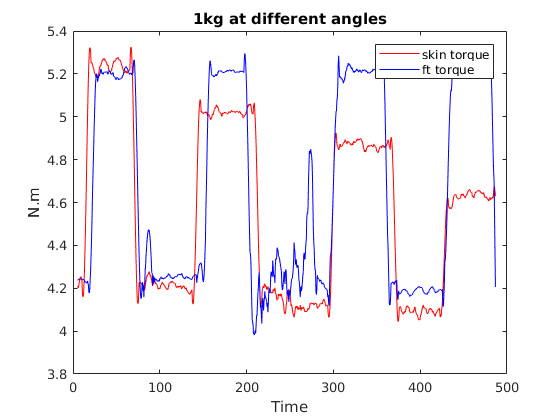}
        \caption{Adding 1kg with the angles: 90, 85, 80, 75} 
        \label{fig:angles}
    \end{subfigure}
\end{figure*}
\label{fig:torqueComparison}

\rowcolors{1}{lightgray}{white}
\begin{table}
 \centering
 \caption{Force and joint torques results}
 \begin{tabular}{|c|c|c|c|c|c|c|} 
  \hline  
  & \multicolumn{3}{|c|}{Forces $N$}   & \multicolumn{2}{|c|}{Joint torques $N.m$}  \\ 
 \hline  
 Masses & Ref. & Simple  & Interpolated & F/T & Interpolated\\ 
 \hline  
 500gr  & 4.905  & 4.6376  & 5.2482 & .  & . \\  
 \hline  
 1000gr & 9.81  & 7.591  & 9.935 & 4.791  & 4.95  \\
 \hline  
 1000gr *& 9.81  & 8.184  & 10.095 & 4.42  & 4.272  \\  
 \hline  
 1500gr & 14.715  & 11.606  & 14.142 & 4.48  & 4.337   \\ 
 \hline  
 1700gr & 16.667  & 14.659  & 17.723 & 4.507  & 4.354 \\ 
 \hline  
 1900gr & 18.639  & 15.528  & 19.054 & 4.529  & 4.374   \\ 
 \hline  
 3000gr* & 29.43  & 22.555  & 27.697 & 4.369  & 4.503   \\ 
 \end{tabular}\\
 *the mass is hanging from the right lower leg
\label{tab:1}
 \end{table}
 
\begin{table}[ht]
\centering
\caption{ Joint torque comparison at different contact angles }

\begin{tabular}{|c|c|c|}
\hline 
 Degree & F/T sensor & Interpolated  \\ 
 90$^o$ & 5.204 & 5.234\\
 85$^o$ & 5.205 & 5.021\\
 80$^o$ & 5.188 & 4.886\\
 75$^o$ & 5.204 & 4.627\\
\hline 
\end{tabular}
\label{tab:2}
\end{table}

 \begin{figure}[ht!]
         \centering
        \includegraphics[width=.45\textwidth]{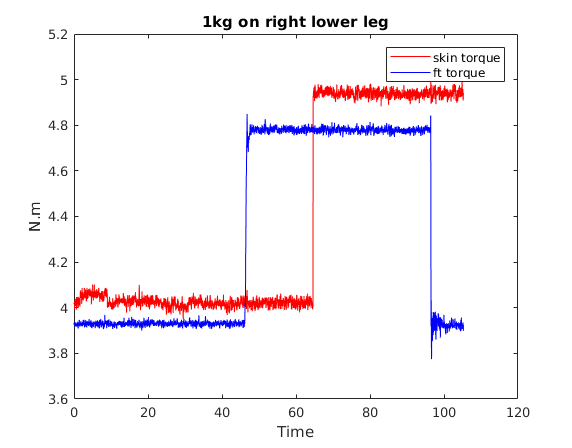}
        \caption{Adding 1kg mass on top of right lower leg}
        \label{fig:noOverlap}
 \end{figure}
 
\subsubsection{Joint torque results}

The current implementation of the interpolation is slow. There is a delay in which the robot actually perceives the change as it can be seen in Fig.\ref{fig:noOverlap}. This is due to the fact that this implementation uses Octave to send the interpolated force-torque, which considering the amount of taxels it measures, can be computationally heavy. The rest of the graphs have been shifted in time to best showcase the comparison between joint torque estimations.

The difference between the F/T estimated torques is around 0.15 N.m on average as it can be verified in Fig. \ref{fig:many}, in some cases it reached even 0.03N.m difference as can be seen from tables \ref{tab:1} and \ref{tab:2}. Considering the the F/T sensors have been effectively used as joint torque feedback for the current controller, these results allow us to consider the joint torques estimated with the skin as viable candidate to replace the F/T measurements.

Considering that real contacts with the environment might not have a constant force, due to the movement either of the robot or the object in contact, Fig. \ref{fig:many} demonstrates how the estimation would respond to slight variations of the contact forces.

When the pressure on the taxels comes close to the 50kPa limit of the calibration the performance dropped, although this can be avoided by distributing the forces over a bigger set of taxels. This allows to correctly estimate cases where it otherwise would not be possible. \ref{fig:3kTorque}.

An important thing to consider though is that the skin measures only normal forces and this effect can be showcased in Fig. \ref{fig:angles} and table \ref{tab:2}. Where the performance of the joint torques estimated with the skin drop due to the angle of the external force.

\section{CONCLUSIONS}
\label{sec:conclusions}

Results have shown that the skin has the possibility to serve as an external force sensor and potentially replace the force/torque sensors. Using the skin is an interesting solution to the joint torque estimation problem because the cost of the distributed tactile skin is lower than the set of ft sensors and easier to integrate. It also allows to correctly estimate external forces when more than one external force is acting on each subtree, which was a limitation of the previous estimation scheme.

While the experiments show results comparable to the F/T sensors, using the iCub skin has the following limitations:

\begin{itemize}
\item Can not measure the sheer forces only normal forces at he contact. 
\item Unable to detect pure torques, or forces aligned with the surface of contact.
\item The pressure in any given taxel should not exceed the max pressure used in the calibration.
\item It is unable to deal with temperature drift.
\end{itemize}

Both a device to surpass the pressure limit in calibration as well as a skin capable of sensing sheer forces are open research topics at the moment.

 It will be interesting to see how the controller responds to this joint torque estimation once we can speed the flow of information into the estimation scheme.
 
Part of the future work is to fuse the information of the F/T sensors, the skin and possibly the motor current to improve the estimation of external force-torques and joint torques, beyond the current results.


\bibliographystyle{IEEEtran}
\bibliography{Literature/bib}

\addtolength{\textheight}{-12cm}

\end{document}